\documentclass{article}

\PassOptionsToPackage{numbers, compress}{natbib}

\usepackage[preprint]{neurips_2025}

\usepackage[utf8]{inputenc} 
\usepackage[T1]{fontenc}    
\usepackage{hyperref}       
\usepackage{url}            
\usepackage{booktabs}       
\usepackage{amsfonts}       
\usepackage{nicefrac}       
\usepackage{microtype}      
\usepackage{subcaption}
\usepackage{graphicx}
\usepackage{algorithm}
\usepackage{algpseudocode}
\usepackage{amsmath}
\usepackage{caption}
\usepackage[table,xcdraw]{xcolor}

\title{Learn from the Past: Fast Sparse Indexing for Large Language Model Decoding}

\author{%
  Feiyu Yao\\
  Beijing University of Technology\\
  \texttt{yaofeiyu10@emails.bjut.edu.cn} \\
  \And
  Qian Wang \\
  Beijing University of Technology \\
  \texttt{wangqian2020@bjut.edu.cn} \\
}

\begin{document}

	\maketitle

	\begin{abstract}
		As large language models (LLMs) continue to support increasingly longer contexts, the memory demand for key-value (KV) caches during decoding grows rapidly, becoming a critical bottleneck in both GPU memory capacity and PCIe bandwidth. Sparse attention mechanisms alleviate this issue by computing attention weights only for selected key-value pairs. However, their indexing computation typically requires traversing all key vectors, resulting in significant computational and data transfer overhead. To reduce the cost of index retrieval, existing methods often treat each decoding step as an independent process, failing to exploit the temporal correlations embedded in historical decoding information. To this end, we propose LFPS(Learn From the Past for Sparse Indexing), an acceleration method that dynamically constructs sparse indexing candidates based on historical attention patterns. LFPS captures two prevalent trends in decoder attention—vertical patterns (attending to fixed positions) and slash patterns (attending to relative positions)—and incorporates a positional expansion strategy to effectively predict the Top-k indices for the current step. We validate LFPS on challenging long-context benchmarks such as LongBench-RULER, using Llama-3.1-8B-Instruct as the base model. Experimental results show that LFPS achieves up to 22.8$\times$ speedup over full attention and 9.6$\times$ speedup over exact Top-k retrieval on an RTX 4090 GPU and a single CPU core of a Xeon Gold 6430, respectively, while preserving generation accuracy. These results demonstrate that LFPS offers a practical and efficient solution for decoding optimization in long-context LLM inference.
		
	\end{abstract}

	\section{Introduction}
	
	With the rapid advancement of large language model technologies, the landscape of applications\cite{achiam2023gpt}\cite{liu2024deepseek} across various industries is undergoing significant transformation. In the inference workflow of LLMs, the process is typically divided into two main stages: the prefill stage and the decoding stage. The prefill stage is characterized by high computational intensity, whereas the decoding stage is predominantly memory-intensive. To improve throughput during decoding, batching techniques\cite{yu2022orca} are commonly employed for greater efficiency. 
	
	As LLMs support longer context lengths\cite{qwen2.5}, their KV caches grow significantly—often exceeding the model’s own VRAM usage. For example, the 16-bit Llama-3-8B model\cite{touvron2023llama}\cite{roziere2023code} uses 16GB of VRAM, while a 128k-token request needs an additional 16GB for KV cache. This growing memory demand during decoding makes GPU memory a major bottleneck. Although systems like vLLM\cite{kwon2023efficient} use paged KV cache management to reduce fragmentation, they do not address the root issue. Offloading KV caches\cite{sheng2023flexgen} to host memory remains an effective solution for multi-batch, long-context inference.
	
\begin{figure}[tbp]
		\centering
		\begin{subfigure}[b]{0.3\textwidth}
			\includegraphics[width=\textwidth]{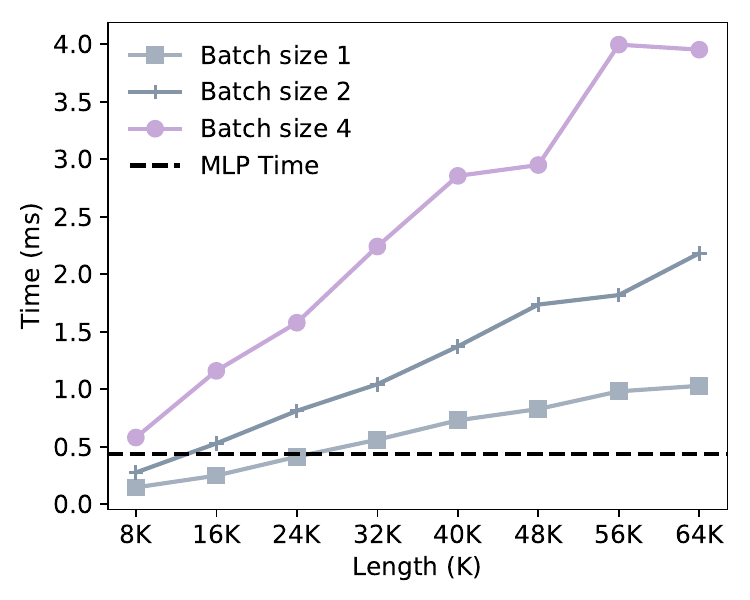}
			\caption{Time to transfer the key}
			\label{fig:1a}
		\end{subfigure}
		\hfill
		\begin{subfigure}[b]{0.3\textwidth}
			\includegraphics[width=\textwidth]{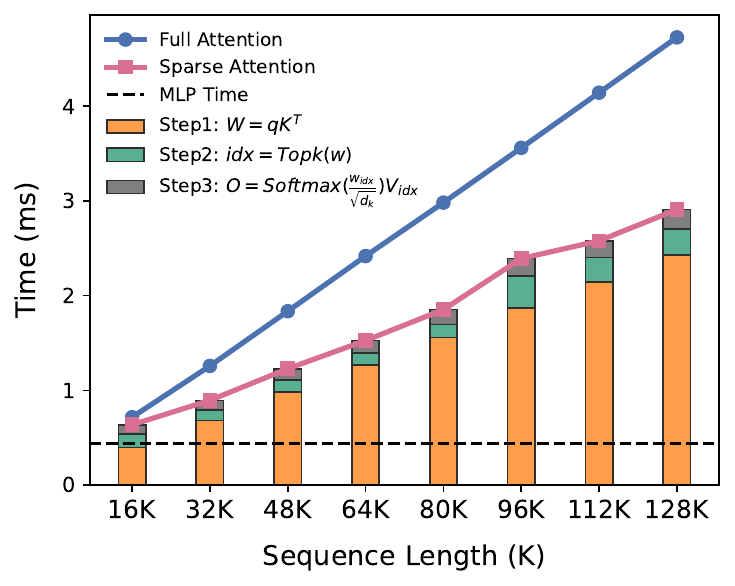}
			\caption{Time to compute sparse index}
			\label{fig:1b}
		\end{subfigure}
		\hfill
		\begin{subfigure}[b]{0.3\textwidth}
			\includegraphics[width=\textwidth]{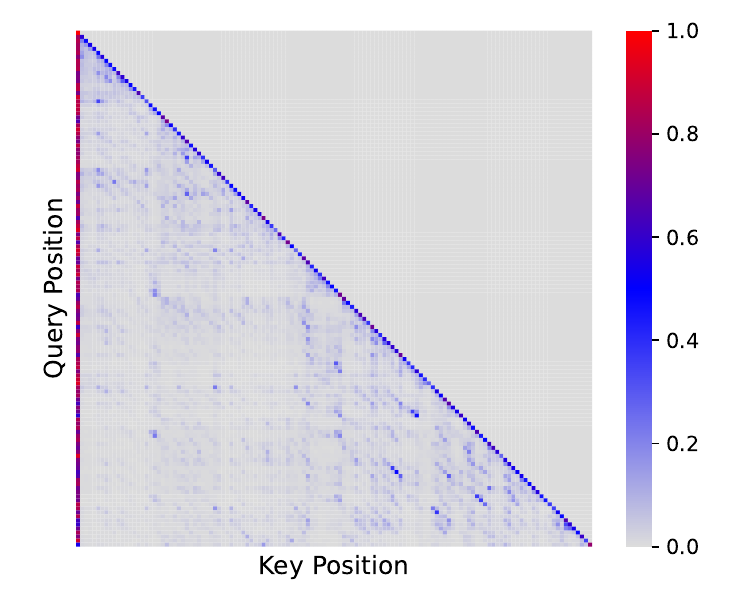}
			\caption{Attention sparsity}
			\label{fig:1c}
		\end{subfigure}
		\caption{(a) At a PCIe bandwidth of 32 GB/s, the time required to transfer 25\% of the key vectors increases with both batch size and context length. (b) The time required for the CPU to compute Top-k (5\%) indices increases with context length, with the batch size fixed at 4. (c) Sparsity of attention. Two common sparsity patterns are observed in attention: vertical and slash. The figure illustrates an example of the slash pattern.}
		\label{fig:1}
	\end{figure}

	KV cache often stores a large number of KV pairs from multiple requests, and in certain scenarios such as multi-turn dialogues, the memory for these KV pairs may not be released even after the request is completed. In the decoding stage, particularly under batch processing scenarios, PCIe bandwidth has emerged as a significant new bottleneck. An effective approach to address this issue is to use sparse attention\cite{zhang2023h2o}\cite{li2024snapkv}, where only a small subset of important KV pairs are weighted in self-attention, enabling the extraction of most of the information while reducing the amount of data transferred over PCIe\cite{jiang2024efficient}. Unfortunately, computing sparse indices precisely requires inner products between the query vector and all key vectors. We have identified two common modes for sparse indexing calculations, each with its own set of issues, as outlined below:
	
	\textbf{Prefetching the compressed key vertor into VRAM for approximate sparse indexing calculations.} However, we demonstrate that although only a small portion of the KV cache is prefetched, due to the memory-intensive nature of the decoding phase, it is often difficult to achieve effective overlap. As shown in Figure \ref{fig:1a}, when only 25\% of the key data from a single layer is transferred\cite{lee2024infinigen}, with PCIe bandwidth at 32GB/s, it is evident that the transmission time increases rapidly as both the KV cache length and batch size increase. In contrast, during the decoding phase, the MLP computation does not scale with the KV cache length but increases gradually with batch size. Thus, prefetching is often unable to overlap effectively with the MLP computation, resulting in limited performance gains.

	\textbf{Offload the query vector to host memory and perform approximate sparse indexing on the CPU.} Because each decoding step handles a query of length 1, its PCIe transfer time is negligible; however, since CPU throughput and parallelism are lower than GPU, as batch size and KV cache length grow, the CPU-side indexing workload can increase rapidly.
	
	To address the PCIe bandwidth bottleneck in the second scenario, prior studies\cite{chen2024magicpig}\cite{he2024fastdecode} have proposed not only offloading the KV cache to host memory but also offloading the attention computation to the CPU. Given that the computational workload during the decoding stage is significantly lighter than that of the prefill stage, these approaches exploit the sparsity of attention to substantially reduce the computational overhead on the CPU. We decompose the sparse attention computation into three steps: \textbf{Step 1:} Transfer the query vector to host memory and compute attention weights $w = qK^T$; \textbf{Step 2:} Selection on $w$ to obtain the sparse indices $idx=TopK(w)$; \textbf{Step 3:} Compute the final output using the sparse indices $O = Softmax(\frac{w_{idx}}{\sqrt{d_k}})V_{idx} $. In Figure \ref{fig:1b}, the computation of sparse indices—namely Step 1—requires a dot product between the query vector and the full set of key vectors. This operation incurs a computational cost that scales linearly with the KV cache length and thus constitutes the primary performance bottleneck. In contrast, Steps 2 and 3 involve only lightweight operations over a small subset of positions and remain efficient even when executed on the CPU, making them negligible in terms of overall performance impact.
	
	Given that the KV cache is stored in host memory and the attention mechanism inherently exhibits strong sparsity, performing sparse index computation on the CPU is a reasonable and efficient choice. As previously discussed, numerous optimization\cite{zhang2024pqcache}\cite{liu2024retrievalattention}\cite{tang2024quest}\cite{zhang2024pqcache} techniques have been proposed to reduce the computational overhead of sparse indexing on the CPU. However, most of these methods treat each decoding iteration as an independent sparse retrieval task, failing to fully exploit the potential correlations among retrieval results across consecutive decoding steps. To address this limitation, we propose a novel sparse indexing acceleration method—LFPS (Learn From the Past for Sparse Indexing). LFPS leverages historical query results from previous decoding steps to efficiently predict the Top-k sparse indices for the current step. By effectively reusing information from earlier decoding iterations, LFPS significantly reduces redundant computations during sparse indexing, thereby improving overall inference efficiency in the decoding stage.
	
	\section{Related work}
	
	\textbf{Based on positional indexing} methods typically rely on predefined positional patterns to select the relevant KV pairs. For instance, StreamingLLM\cite{xiao2023efficient} has observed the "attention sink" phenomenon, where only the KV pairs at the beginning of the sequence and the most recent ones are retained. DuoAttention\cite{xiao2024duoattention}, on the other hand, determines whether each attention head is a Streaming Head to decide whether to adopt a sparse attention mechanism. However, relying solely on fixed positional patterns makes it difficult to accurately select the most critical KV pairs for the current decoding step, leading to low retrieval accuracy, consequently, affecting overall inference performance.
	
\begin{figure}[htbp]
		\centering
		\begin{subfigure}[b]{0.3\textwidth}
			\includegraphics[width=\textwidth]{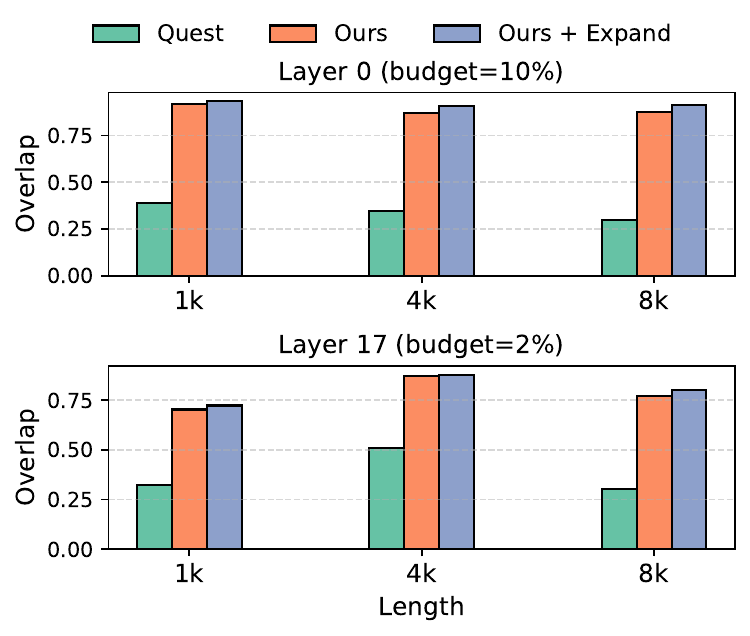}
			\caption{The overlap with TopK}
			\label{fig:2a}
		\end{subfigure}
		\hfill
		\begin{subfigure}[b]{0.3\textwidth}
			\includegraphics[width=\textwidth]{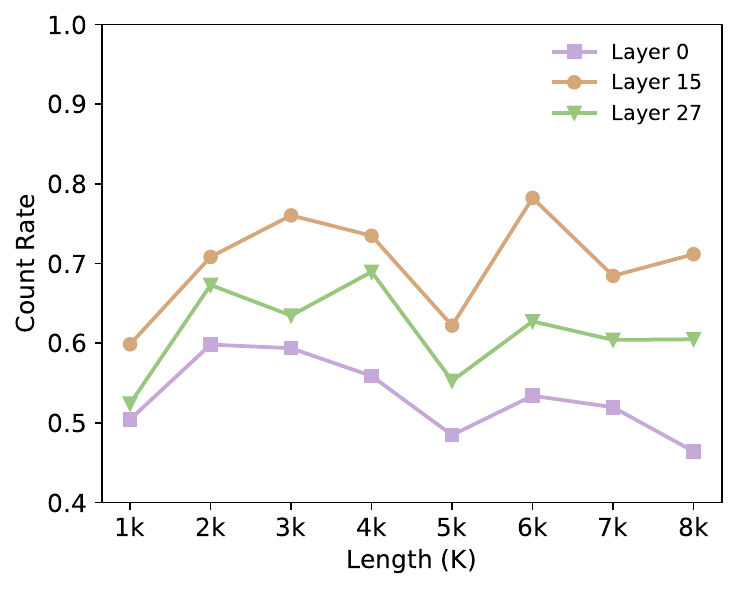}
			\caption{Important KV pairs are closer.}
			\label{fig:2b}
		\end{subfigure}
		\hfill
		\begin{subfigure}[b]{0.3\textwidth}
			\includegraphics[width=\textwidth]{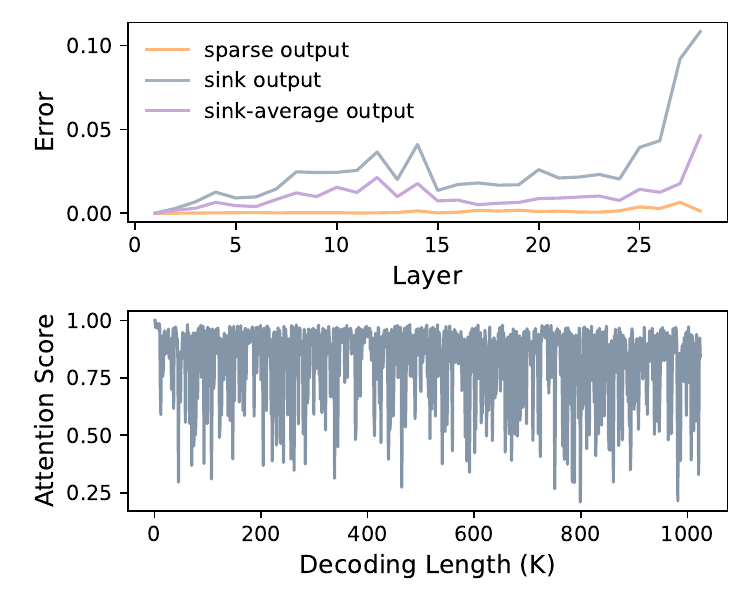}
			\caption{The error with full Attention}
			\label{fig:2c}
		\end{subfigure}
		\caption{(a) Across various sequence lengths and budget settings, our method achieves a higher overlap with the exact Top-k sparse selection compared to Quest. (b) Under a 2\% budget, approximately 60\% of important KV pairs are within a distance of 2 from each other. (c) Error relative to Full Attention: applying mean aggregation to unselected KV pairs results in lower approximation error. Top: Error analysis under high sparsity conditions. Sparse output denotes the result of exact Top-k sparse attention; sink output refers to using only the sink token for weighted output; sink-average output combines the sink token with the mean of other KV pairs. Bottom: Attention scores of the sink token during decoding of tokens 0–1k in head 0 of layer 23. Visualizations are based on Llama-3.1-8B-Instruct evaluated on a single RTX 4090 GPU.}
		\label{fig:2}
	\end{figure}

	\textbf{Block-based indexing} methods organize KV pairs into blocks\cite{liu2024deepseek}\cite{lu2025moba} and perform indexing calculations at the block level. In this approach, Quest\cite{tang2024quest} and InfLLM\cite{xiao2024infllm} use fixed, contiguous KV pairs to form blocks, while PQ Cache\cite{zhang2024pqcache} and Cluster KV\cite{liu2024clusterkv} partition blocks by reclustering, and EM-LLM\cite{fountas2024human} can automatically determine block boundaries. Although indexing at the block level helps control the cost of sparse indexing selection to some extent, not all KV pairs within a block are equally important, leading to budget waste. Figure \ref{fig:2a} illustrates the overlap rate between the sparse indexing selection of Quest and the precise sparse indexing selection, with the overlap rate falling below 50\%. Furthermore, for extremely long contexts, indexing at the block level still struggles to significantly reduce the cost of indexing calculations.

	\textbf{Low-rank.} Numerous studies have shown that the Query and Key matrices in KV cache exhibit low-rank characteristics. To address this, these methods perform low-rank projections on the Query and Key matrices to reduce the computational cost of sparse indexing. Techniques such as InfiniGen\cite{lee2024infinigen}, ShadowKV\cite{sun2024shadowkv}, Loki\cite{bassil2017loki}, and SparQ\cite{ribar2023sparq} have proposed efficient low-rank compression methods that can significantly accelerate the sparse indexing process, and these approaches are orthogonal to ours.

	\textbf{Sampling.} MagicPIG\cite{chen2024magicpig} demonstrates that by sampling and selecting the most important KV pairs and weighting them, it can achieve smaller errors than Topk. To reduce the sampling overhead, MagicPIG introduces an approximate sampling method based on Local Sensitive Hashing (LSH), significantly reducing the computational cost of sampling probabilities. However, LSH introduces biases when estimating sampling probabilities, and these biases are highly dependent on the parameter configuration. Furthermore, the memory consumption of the LSH hash table is nearly identical to that of the full KV cache, making it challenging to deploy the system on resource-constrained devices, especially in high-throughput inference scenarios.
	
	\section{Motivation}
	
	As mentioned earlier, these methods typically treat each decoding iteration as an independent sparse indexing retrieval task. This raises an important question: under the premise of ensuring generation accuracy, can we, using tokens as the basic unit, fully leverage the retrieval results from previous decoding steps to significantly reduce the retrieval overhead of the current decoding step? The answer is yes. Our method is based on the following three key findings(See Appendix~\ref{app:Supplementary Analysis} for more details).
	
	\textbf{1) In attention maps, two typical patterns commonly emerge: vertical and slash\cite{jiang2024minference}.} The vertical pattern reflects the model's focus on the importance of certain fixed KV pairs, while the slash pattern indicates the model's attention to fixed relative positions. Figure \ref{fig:1c} shows a slash attention map. Additionally, related research\cite{wang2024model} has shown that neighboring Key vectors and Query vectors typically exhibit high cosine similarity, suggesting that retrieval results across different decoding steps are significantly similar. Based on these observations, we collect and continuously update the attention scores for both vertical and slash patterns. This allows us to pre-filter some unimportant KV pairs before generating the Query vector, thereby constructing an initial index candidate set. Retrieval at the token level shows a high overlap rate— as shown in Figure \ref{fig:2a}, with an overlap rate of 85\%. This phenomenon further indicates that leveraging historical retrieval results can, to some extent, reduce the retrieval cost in the current decoding process.
	
	\textbf{2) Important KV pairs often exhibit a clustering effect, where they tend to be sparsely distributed across various positions after a small amount of aggregation.} This clustering effect does not imply that important KV pairs form contiguous blocks, but rather that their relative distances are relatively short. As shown in Figure \ref{fig:2b}, when retaining only 2\% of the budget, the proportion of important KV pairs with a distance of 2 or less is between 50-80\%. Therefore, based on the initial index candidate set, we introduce a position expansion strategy to construct an expanded index candidate set. This strategy, based on local positions, includes KV pairs adjacent to the initial important KV pairs into the candidate range, further improving the retrieval coverage. As shown in Figure \ref{fig:2a}, the overlap rate increases by approximately 3\% after expansion. More importantly, position expansion can capture potential important KV pairs that were not identified in the original vertical and slash attention patterns, making the attention score updates for both patterns more robust.
	
	\textbf{3) Highly sparse attention heads do not require self-attention weighting, and the error is smaller when the KV pairs outside of the sparse indexing are averaged.} When an attention head exhibits high sparsity, its attention weights are often concentrated on the sink token, indicating that the head has failed to effectively focus on the key information in the sequence. The weighted output of this head is essentially equivalent to the value vector corresponding to the sink token. Therefore, we can conclude that: 1) For highly sparse attention heads, the value vector of the sink token can be directly returned as the output, bypassing redundant weighting calculations; 2) As shown in Figure \ref{fig:2c} (upper part), compared to returning only the value of the sink token, using the KV pair of the sink token along with the mean of other non-sparse KV pairs to form a weighted result leads to a smaller error with the full attention output. 3) In such heads, attention weights outside of the sink token tend to be more evenly distributed, resulting in smoother attention scores for vertical and slash patterns, thereby weakening the distinguishability and significance of these patterns. For example, as shown in Figure \ref{fig:2c} (lower part), during the decoding process, the weight of the sink token in certain attention heads exhibits significant oscillation between high and low sparsity states. This high sparsity introduces noticeable noise in the attention score updates based on the vertical and slash patterns.

	In summary, by leveraging the three key points mentioned above, we can significantly reduce the indexing overhead in processing long contexts while ensuring accuracy. We optimize computational efficiency by offloading the query vectors to host memory and transferring the sparse indexing and weighted result calculations to the CPU. Although there is a notable difference in computational power between the CPU and GPU, the substantial reduction in sparse indexing calculations leads to a significant decrease in overall computational demand, effectively reducing the latency caused by data transmission via PCIe. Furthermore, existing research\cite{he2024fastdecode}\cite{jiang2024neo}\cite{jin2024compute} has proposed dividing different batches and attention heads into smaller sub-batches to achieve a computational load balance between the GPU and CPU, further enhancing system computational efficiency and resource utilization.

	\section{Method}

\begin{figure}[htbp]
		\centering
		\includegraphics[width=\textwidth]{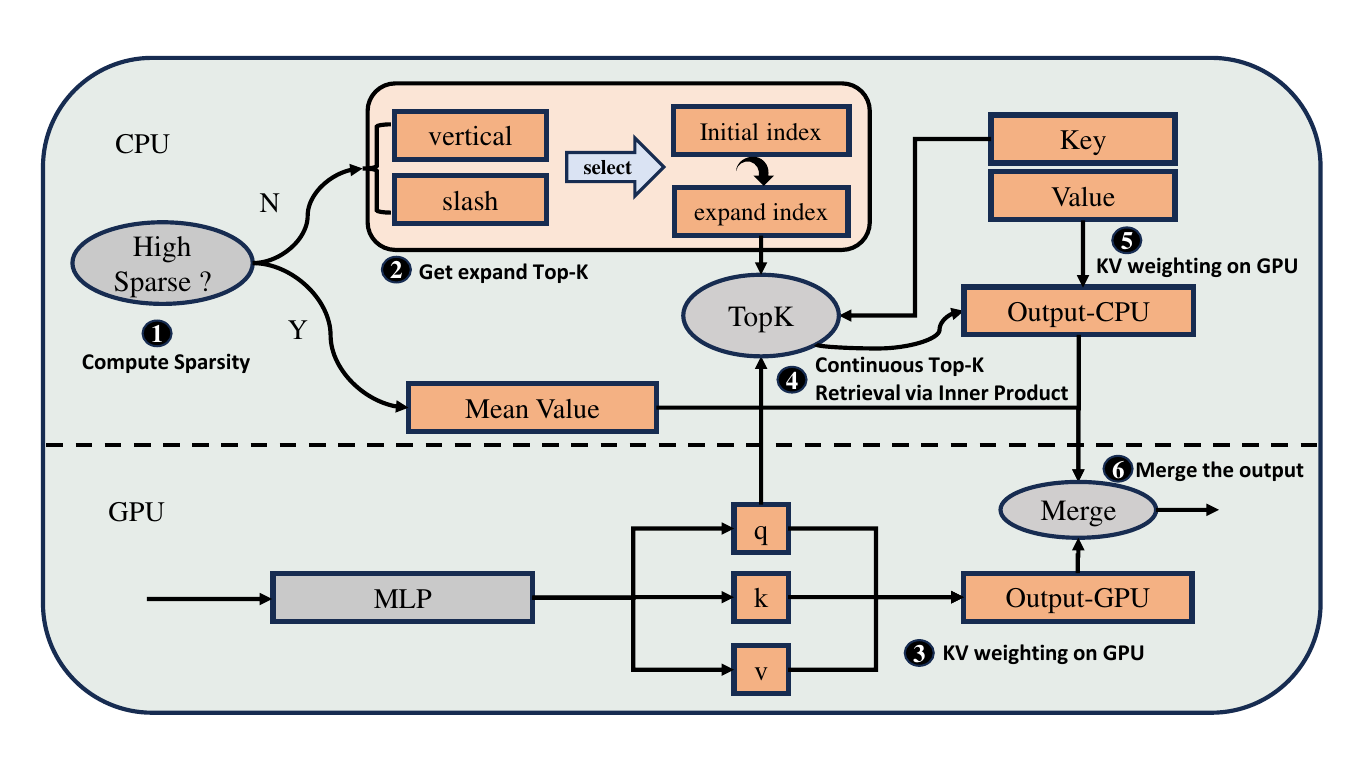}
		\caption{The overview of LFPS.}
		\label{fig:3}
	\end{figure}

	In this section, we introduce LFPS, a sparse indexing acceleration method based on historical decoding patterns. The core innovation lies in dynamically constructing candidate sets through the reuse of sparse attention patterns generated during historical decoding processes, thereby reducing the computational complexity of Top-k operations in current decoding steps. As illustrated in Figure \ref{fig:3}, the LFPS framework operates through two phases.
	
	During the prefilling phase, the system offloads the model's KV cache to host memory while extracting two characteristic attention patterns: vertical and slash. In the decoding phase, prior to generating the current query vector, the system leverages pre-collected attention patterns to facilitate rapid retrieval of extended indices. Subsequent processing involves executing inner-product operations exclusively on these extended indices to achieve precise index filtering and weighted aggregation of value vectors. Notably, for attention heads exhibiting high sparsity characteristics, the system implements a computational bypass strategy that substitutes real-time weighted calculations with mean value vectors, effectively mitigating redundant computations while minimizing interference with the established sparse attention patterns during pattern updates.

	\subsection{Problem Formulation}
	
	During the decoding phase of large language models, given the current query vector at step $t$, denoted as $q^{(t)} \in \mathbb{R}^{d}$, and the key matrix $K^{(t)} \in \mathbb{R}^{n \times d}$ stored in host memory—where $n$ represents the context length and $d$ the dimensionality of the vectors. The objective of sparse indexing is to efficiently identify the Top-k most relevant indices $\mathcal{I}^{(t)}$ from $K^{(t)}$. These selected indices are then used to compute the weighted output.
	\begin{equation}
		O(\mathcal{I}^{(t)}) = \mathrm{Softmax}\left( \frac{q^{(t)} {K_{\mathcal{I}}^{(t)}}^\top}{\sqrt{d}} \right) V^{(t)}_{\mathcal{I}}
		\label{eq:1}
	\end{equation}
	where $V^{(t)} \in \mathbb{R}^{n \times d}$ denotes the value matrix, and $K_\mathcal{I}^{(t)}$, $V_\mathcal{I}^{(t)}$ represent the key and value vectors corresponding to the selected index set $\mathcal{I}^{(t)}$, respectively.
	Let the selected set $\mathcal{C}^{(t)} \subseteq \{0, 1, 2, \dots, n-1\}$ be an approximate subset of the true Top-k indices $\mathcal{I}^{(t)}$. The overlap ratio is defined as:
	\begin{equation}
		\eta = \frac{|\mathcal{C}^{(t)} \cap \mathcal{I}^{(t)}|}{k} \in [0, 1]
		\label{eq:2}
	\end{equation}
	The LFPS method is designed to minimize the computational cost of sparse indexing by dynamically constructing the candidate set $\mathcal{C}^{(t)}$ through the reuse of historical decoding information. Its objectives can be formalized as follows:
	\begin{equation}
		\min \ \mathbb{E}[|\mathcal{C}^{(t)}|]   \quad \max \ \mathbb{E}\left[\frac{|\mathcal{C}^{(t)} \cap \mathcal{I}^{(t)}|}{k}\right]
		\label{eq:3}
	\end{equation}
	
	\subsection{The LFPS algorithm}
	
	\textbf{Prefilling.} Given an input sequence $X \in \mathbb{R}^{n \times d}$, and the projected key and value matrices defined as $K = XW_K$, $V = XW_V$, where $W_K$ and $W_V$ are projection matrices, the computed results are offloaded to host memory. For each attention head, based on the attention weights from the prefill stage, we construct an initial score table \footnote{We retain the sink token on the GPU to prevent bias during the construction of the initial score table. In the following discussion, we focus exclusively on the offloaded KV cache.} using only the attention weights from the most recent $s$ decoding steps, denoted as $\{w^{(t)}\}_{t=n-s}^{n-1}$.
	\begin{equation}
		\Phi _{ver}^{(0)}(i) = \frac{1}{2s(1-r)} \sum_{j=1}^{s} w^{(n-j)}_i \quad \Phi _{sla}^{(0)}(i) = \frac{1}{2s(1-r)} \sum_{j=1}^{s} w^{(n-j)}_{i-j+1}
		\label{eq:4}
	\end{equation}
	where $ w^{t}_i $ represents the attention score at position $i$ during the 
	$t$-th decoding step, with the boundary condition $ w_j(i-j+1) = 0 $ when $ i-j+1 < 0 $. The factor $ \frac{1}{2s(1 - r)} $ is the normalization coefficient, where $ r $ is the decay factor used in the score table update.
	
	In addition, we store, for each head $h$ during the prefill stage, the mean key and value vectors $\bar{K}_h$ and $\bar{V}_h$, respectively, as well as the ratio of the variance $\sigma_h^2$ to the squared norm of the corresponding query $\|\mathbf{q}_h\|^2$, as global semantic priors that can be reused during decoding: $\hat{\sigma}_h^2 = \frac{\sigma_h^2}{\|\mathbf{q}_h\|^2} $ (Use the final decoding step from the prefill stage for computation).

	\begin{figure}[H]
		\centering
		\begin{minipage}{0.48\textwidth}
			\begin{algorithm}[H]
				\caption{LFPS Decoding Step}
				\label{alg:lfps-decoding}
				\begin{algorithmic}[1]
					\Require Query vector $q^{(t)}$, key matrix $K^{(t)}$, score tables $\Phi_{\mathrm{ver}}^{(t-1)}$, $\Phi_{\mathrm{sla}}^{(t-1)}$
					\Ensure Top-k index set $\mathcal{I}^{(t)}$
					\State Compute sparsity ratio $\rho_h^{(t)}$ using Eq.~\eqref{eq:6}
					\If{$\rho_h^{(t)} > \varepsilon$}
					\State \Return $\bar{V}_h$
					\EndIf
					\State Compute thresholds $\tau_{\mathrm{ver}}$, $\tau_{\mathrm{sla}}$ using Eq.~\eqref{eq:8}
					\State Select initial candidate index $\mathcal{C}_0^{(t)}$ by Eq.~\eqref{eq:7}
					\State Expand to $\mathcal{C}_1^{(t)}$ from $\mathcal{C}_0^{(t)}$ by Eq.~\eqref{eq:9}
					\State Select Top-k indices $\mathcal{I}^{(t)}$ via exact dot-product over $\mathcal{C}_1^{(t)}$
					\State Update score tables using Eq.~\eqref{eq:10}
					\State \Return $\mathcal{I}^{(t)}$
				\end{algorithmic}
			\end{algorithm}
		\end{minipage}
		\hfill
		\begin{minipage}{0.48\textwidth} 
			\includegraphics[width=\linewidth]{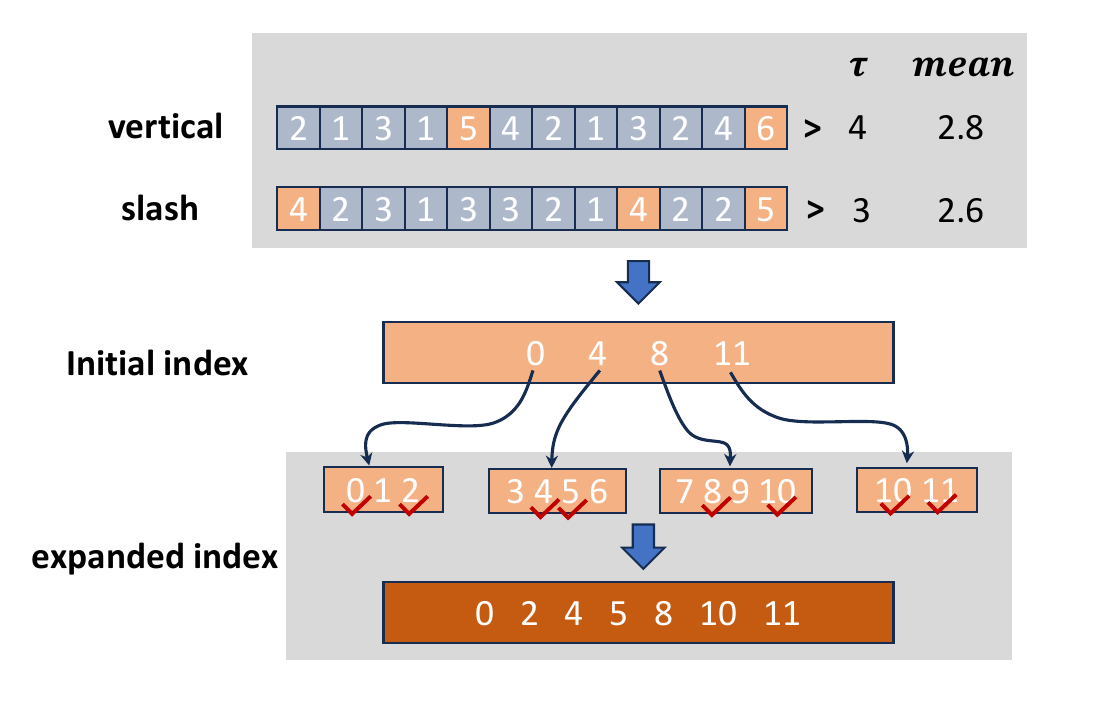} 
			\captionof{figure}{For example, given the vertical and slash score table, we first compute their thresholds and means using Equation~\eqref{eq:8}. The initial candidate index is obtained based on the thresholds and then expanded with offsets 
				\{-1,0,1,2\}, keeping only indices whose scores exceed the corresponding mean.}
			\label{fig:4}
		\end{minipage}
	\end{figure}
	
	\textbf{Decoding.} At each decoding step $t$, given the query vector $q^{(t)}\in\mathbb{R}^d$, the key matrix $K^{(t)}\in\mathbb{R}^{n\times d}$, and the score tables $\Phi_{ver}^{(t-1)},\Phi_{sla}^{(t-1)}$, the complete decoding process is depicted in Figure \ref{fig:4} and summarized in Algorithm \ref{alg:lfps-decoding}. LFPS dynamically constructs the candidate index set through four stages:
	
	\textbf{1) Head sparsity estimation and sink-value bypass.} At decoding step $t$, let the attention weights for the head $h$ be $w^{(t)} = [w^{(t)}_0, \dots, w^{(t)}_{n-1}]$.  We propose three key components for sparsity characterization(More details are provided in Appendix~\ref{app:Head sparsity estimation}):
	\begin{align}
		w^{(t)}_{\text{sink}} &= \exp\left( \frac{q^{(t)} K_{\text{sink}}}{\sqrt{d}} \right) \notag \\
		w^{(t)}_{\text{global}} &= \exp\left( \frac{q^{(t)} \bar{K}_h}{\sqrt{d}} + \frac{\|q^{(t)}\|^2 \cdot \hat{\sigma}_h^2}{2} \right) \cdot n \notag \\
		w^{(t)}_{\text{local}} &= \sum_{i=t-6}^{t-1} \exp\left( \frac{q^{(t)} K_i}{\sqrt{d}} \right)
		\label{eq:5}
	\end{align}
	
	The head-specific sparsity ratio is computed as: 
	\begin{equation}
		\rho_h^{(t)} = \frac{w^{(t)}_{sink}}{w^{(t)}_{sink} + w^{(t)}_{global} + w^{(t)}_{local}}
		\label{eq:6}
	\end{equation}
	When the sparsity score $ \rho_h^{(t)} $ exceeds a predefined threshold $ \varepsilon $ (e.g., 0.85), the corresponding head is regarded as highly sparse. In such cases, we skip steps 2), 3), and 4), as highly sparse heads tend to smooth other attention scores and introduce noise during the update in step 4). In contrast, the output is obtained by directly averaging the value vectors $\bar{V}_h$.
	
	\textbf{2) Initial candidate index selection.} We form the initial  candidate index set
	\begin{equation}
		\mathcal{C}_0^{(t)} \;=\; \bigl\{\,i \mid \Phi_{\mathrm{ver}}^{(t)}(i) > \tau_{\mathrm{ver}}\bigr\}
		\;\cup\;
		\bigl\{\,i \mid \Phi_{\mathrm{sla}}^{(t)}(i) > \tau_{\mathrm{sla}}\bigr\},
		\label{eq:7}
	\end{equation}
	where the thresholds $\tau_{\mathrm{ver}}$ and $\tau_{\mathrm{sla}}$ are set dynamically based on the approximate kurtosis of the two score distributions.  Concretely, let
	\begin{equation}
		\kappa(x) = \frac{\sum_{i=1}^{n}(x_i-\bar{x})^4}{(\sum_{i=1}^{n}(x_i-\bar{x})^2)^2} \quad \tau(x) = a\times \frac{\frac{1}{n}\sum_{i=1}^{n} x_i}{\kappa (x)}
		\label{eq:8}
	\end{equation}
	A higher kurtosis indicates a sharper and more pronounced peak in the score distribution, suggesting stronger pattern significance, which triggers the system to automatically allocate a larger index budget. Here, $a$ is hyperparameter that control the threshold. Different values of $a$ control different retrieval budgets, and please refer to~\ref{app:Parameter Selection in LFPS}
	
	\textbf{3) Candidate set expansion.} To include neighboring KV pairs, expand each index in $\mathcal{C}_0^{(t)}$ by offsets \{-1,0,1,2\} and keep those whose vertical or slash importance exceeds its mean value:
	\begin{equation}
		\mathcal{C}_1^{(t)}=
		\bigcup_{i\in\mathcal{C}_0^{(t)}}
		\{
		j = i + \delta \bigm|
		\delta\in\{-1,0,1,2\},
		(\Phi_{ver}^{(t)}(j)>\bar{\Phi}^{(t)}_{ver})
		\lor
		\Phi_{\mathrm{sla}}^{(t)}(j)>\bar{\Phi}^{(t)}_{sla})
		\bigr\}
		\label{eq:9}
	\end{equation}
	
	\textbf{4) Update of the vertical and slash score tables.} After obtaining the final Top‑k index set $\mathcal{C}_2^{(t)}$ via exact dot-product evaluation within the expanded index set $\mathcal{C}_1^{(t)}$, we update the vertical and slash score tables for each attention head as follows:
	\begin{equation}
		\Phi_{ver}^{(t+1)}(i) = r\Phi_{ver}^{(t)}(i) + w^{(t)}_i - \frac{1}{2|\mathcal{C}_2^{(t)}|}
		\quad
		\Phi_{sla}^{(t+1)}(i) = r\Phi_{sla}^{(t)}(i-1) + w^{(t)}_i - \frac{1}{2|\mathcal{C}_2^{(t)}|}
		\label{eq:10}
	\end{equation}
	where $r \in [0,1)$ is a decay factor controlling the memory of past scores, and $\frac{1}{2|\mathcal{C}_2^{(t)}|}$ represents half of the normalized attention weight. For $i \notin \mathcal{C}_2^{(t)}$, we define $w^{(t)}_i = \frac{1}{2|\mathcal{C}_2^{(t)}|}$. These updates keep the importance scores stable over time by combining new attention signals with past values.
	
	\section{Experiments}
	
	This section aims to validate the effectiveness and decoding efficiency of the proposed LFPS method in long-context sparse indexing scenarios. We first present the experimental setup, and then compare LFPS against various sparse attention methods across different benchmarks and configurations in terms of both accuracy and efficiency.
	
	\paragraph{Setup.} Our experiments are based on the Llama-3.1-8B-Instruct model. During inference, we retain only the first four tokens of the input sequence and adopt a greedy decoding strategy to ensure the stability of the generated output. We first evaluate the LFPS method on two long-sequence benchmarks LongBench\cite{bai2023longbench} and RULER\cite{hsieh2024ruler} and report the average performance across different task types to comprehensively assess the method’s stability and generalization ability. During decoding, the KV pairs offloaded to host memory are processed on the CPU for attention computation. For latency evaluation, we test the single-core and multi-core performance of our method on two hardware configurations: a 16 vCPU Intel Xeon  Gold 6430 with an RTX 4090, and a 20 vCPU Intel  Xeon Platinum 8457C with an L20 GPU (48GB memory).

	\paragraph{Baseline.} For baselines, full Attention and exact Top-k serve as upper bounds, as they perform precise attention computation over the entire context. StreamingLLM\cite{xiao2023efficient} retains the initial four tokens and a fixed number of KV pairs based on a predefined budget, following a fixed retention pattern. Quest\cite{tang2024quest} selects relevant information at the granularity of blocks, for which we set the block size to 16. MagicPIG\cite{chen2024magicpig} employs locality sensitive hashing (LSH) to sample KV pairs, with LSH parameters configured as $K=10$ and $L=150$. In our method, we set the recent decoding step $s=32$, the decay factor $r=0.95$, the threshold $ \varepsilon = 0.85$ and the adaptation parameter $a$ to 0.2 and 0.3, respectively, for experimentation.
	
	\begin{table}[ht]
		\centering
		\setlength{\tabcolsep}{3pt}
		\caption{Performance of different methods on LongBench tasks under a 2\% attention budget.}
		\begin{tabular}{@{}l|rrrrrrrrr|r@{}}
			\toprule
			Method                      & Qas & HPQ  & 2WQ & MN & TREC  & REQ & SAM & LCC & REP & Avg \\
			\midrule
			Llama-3.1-8B-Instruct & 43.52 & 61.09 & 44.10 & 25.67 & 71.33 & 92.16 & 42.18 & 66.47 & 52.61 & 55.57 \\
			Topk                 & 43.36 & 60.70 & 44.66 & 25.64 & 71.33 & 91.91 & 42.25 & 66.54 & 51.04 & 55.27 \\
			Streaming     & 18.60 & 34.64 & 25.56 & 12.18 & 35.17 & 42.54 & 15.14 & 24.19 & 23.64 & 25.52 \\
			Quest                & 39.73 & 60.01 & 43.02 & 23.93 & 69.67 & 91.76 & \textbf{41.25} & 62.13 & 48.92 & 53.71 \\
			MagicPIG             & 39.09 & 58.34 & 39.57 & \textbf{25.40} & 70.33 & 90.21 & 38.83 & 60.39 & 46.52 & 52.30 \\
			\rowcolor{blue!5}
			LFPS($a=0.2$)     & \textbf{42.01} & \textbf{60.08} & 44.68 & 23.84 & \textbf{71.33} & \textbf{92.21} & 41.23 & \textbf{66.43} & \textbf{50.12} & \textbf{54.66} \\
			\rowcolor{blue!5}
			LFPS($a=0.3$)     & 41.03 & 59.95 & \textbf{45.46} & 23.57 & 71.00 & 92.00 & 41.02 & 66.39 & 50.10 & 54.50 \\
			\bottomrule
		\end{tabular}
		\label{tab:longbench}
	\end{table}
	
	\begin{table}[htbp]
		\centering
		\caption{
			Comparison of different methods on the RULER benchmark under a 2\% attention budget, with evaluations conducted at context lengths from 4K to 128K.
		}
		\begin{tabular}{l|cccccc|c}
			\toprule
			Methods & 4k & 8k & 16k & 32k & 64k & 128k & Avg \\
			\midrule
			Llama-3.1-8B-Instruct & 99.06 & 98.20 & 97.76 & 95.35 & 90.27 & 80.28 & 93.15 \\
			Topk           & 96.96 & 95.88 & 95.60 & 93.48 & 88.56 & 77.94 & 91.07 \\
			Streaming      &  1.47 &  2.00 &  1.88 &  0.82 &  0.02 &  0.12 &  1.05 \\
			Quest          & 77.71 & \textbf{81.40} & 84.45 & 84.90 & \textbf{83.35} & 73.15 & 80.83 \\
			MagicPIG              & 54.40 & 72.84 & 76.07 & 78.21 & 70.35 & 54.40 & 67.05 \\
			\rowcolor{blue!5}
			LFPS($a=0.2$)          & \textbf{77.84} & 81.25 & \textbf{84.82} & \textbf{86.05} & 82.90 & \textbf{73.21} & \textbf{81.01} \\
			\midrule
		\end{tabular}
		
		\label{tab:ruler}
	\end{table}

	\paragraph{LongBench.}  
	As shown in Table~\ref{tab:longbench}, with $a = 0.2$, LFPS’s average performance nearly matches the sparse‐attention upper bound compared to Quest and MagicPIG, while StreamingLLM’s fixed KV‐selection pattern results in significantly lower overall accuracy. Notably, in code‐generation tasks, LFPS exhibits a clear advantage on both the LCC and REP metrics.

	\paragraph{RULER.}  
	In the RULER benchmark, our evaluation spans context lengths ranging from 4K to 128K, aiming to comprehensively assess the accuracy and robustness of various sparse attention methods under extremely long input sequences. Table~\ref{tab:ruler} presents the experimental results of different methods across these context lengths, where LFPS also demonstrates outstanding adaptability to long contexts. Notably, when the context length reaches 128K, LFPS utilizes only 1.7\% of the retrieval budget while still achieving performance slightly surpassing that of Quest (6.25\%).

	\begin{figure}[htbp]
		\centering
		\begin{subfigure}[b]{0.48\textwidth}
			\includegraphics[width=\textwidth]{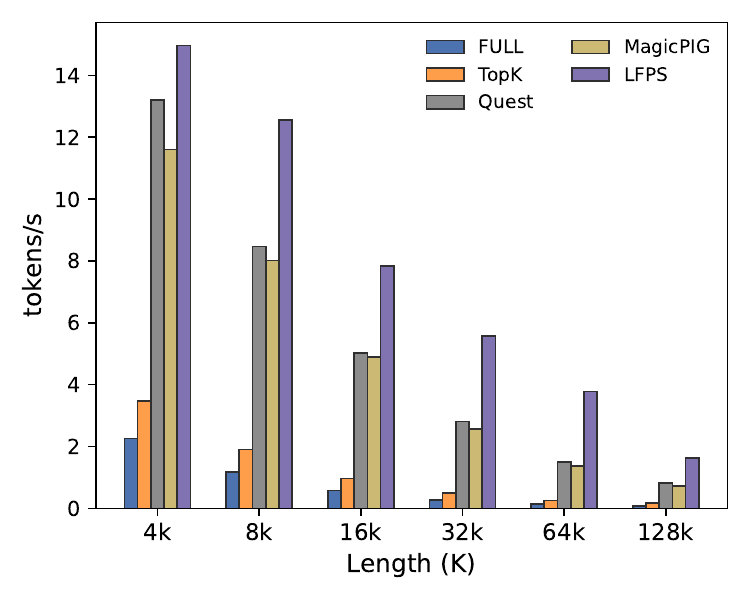}
			\caption{Single-core Performance across Context Lengths}
			\label{fig:5a}
		\end{subfigure}
		\hfill
		\begin{subfigure}[b]{0.48\textwidth}
			\includegraphics[width=\textwidth]{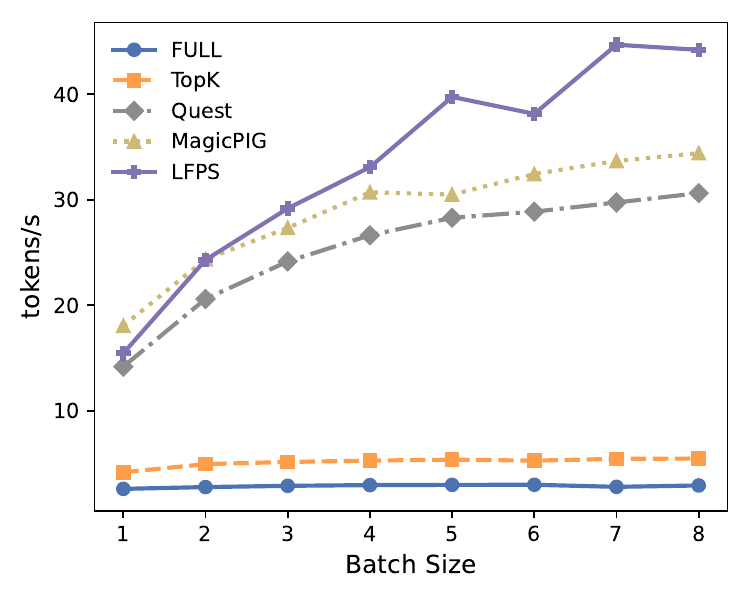}
			\caption{Muti-core Performance across Batch Sizes}
			\label{fig:5b}
		\end{subfigure}
		\caption{
			We evaluate the decoding throughput of Llama-3.1-8B-Instruct with and without LFPS under two hardware configurations. 
			(a) Single-core performance across varying context lengths (4K to 128K tokens) on an NVIDIA RTX 4090 GPU paired with a single-core Intel Xeon Gold 6430 CPU.
			(b) 8-core performance with varying batch sizes (1 to 8) on an NVIDIA L20 GPU coupled with an Intel Xeon Platinum 8457C CPU. 
		}

		\label{fig:5}
	\end{figure}
	
	\paragraph{Latency.} We evaluated the computational efficiency of LFPS under two hardware configurations. Figure \ref{fig:5a}  (single-core, 4K--128K context lengths) illustrates that LFPS maintains a high decoding speed as the context length increases. Specifically, compared to exact Top-k, LFPS achieves speedups of 4.3$\times$, 8.1$\times$, and 14.7$\times$ at context lengths of 4K, 16K, and 64K, respectively. Even at 128K, LFPS is capable of processing at 2 tokens/s, yielding a 22.8$\times$ speedup over full attention and a 2$\times$ speedup over Quest ($<$1 token/s). Figure \ref{fig:5b} (8-core, batch size from 1 to 8) demonstrates that LFPS exhibits significant speed advantages as batch size increases. When the batch size reaches 8, LFPS achieves a token generation rate of 45 tokens/s, representing a 50\% improvement over Quest. This efficiency gain is partially attributed to the reduction in head computations introduced by Equation~\ref{eq:6}, which decreases the number of heads by approximately 12\%.

	\section{Conclusion} 
	This paper addresses the PCIe‐bandwidth bottleneck and reduced computational efficiency that arise when a model’s KV cache is offloaded to host memory during large language model inference. We introduce LFPS, a dynamic sparse‐indexing mechanism that continuously tracks two attention patterns—vertical and slash observed in the decoding history. By constructing a dynamic candidate‐index set based on these patterns, LFPS substantially reduces the number of KV pairs that must be retrieved. Our experiments on an RTX~4090 GPU and a Xeon~Gold~6430 CPU demonstrate that, on a single core, LFPS achieves a $22.8\times$ speedup over full attention and a $9.6\times$ acceleration for exact top–k retrieval. These results indicate that LFPS is a promising approach for accelerating decoding in large language model inference.

	\bibliographystyle{plain}
	\bibliography{references}

	\newpage
	\appendix
	
	\section{Supplementary Analysis}
	\label{app:Supplementary Analysis}
	
	\subsection{The Patterns of Attention}
	
	In the self-attention maps during inference, two predominant patterns are commonly observed: vertical and slash. In Minference\cite{jiang2024minference}, the $\Lambda$-shape and block‑sparse patterns are introduced. The $\Lambda$-shape pattern simultaneously preserves the KV pairs from the beginning of the sequence and those that are relatively close to the current position: the former are easily captured by the vertical pattern, while the latter are easily captured by the slash pattern. The block‑sparse pattern demonstrates a clustering effect on critical KV pairs; by introducing a positional expansion strategy within the candidate set, this pattern can more fully capture such clustering features.
	
\begin{figure}[htbp]
		\centering
		\begin{subfigure}[b]{0.48\textwidth}
			\includegraphics[width=\textwidth]{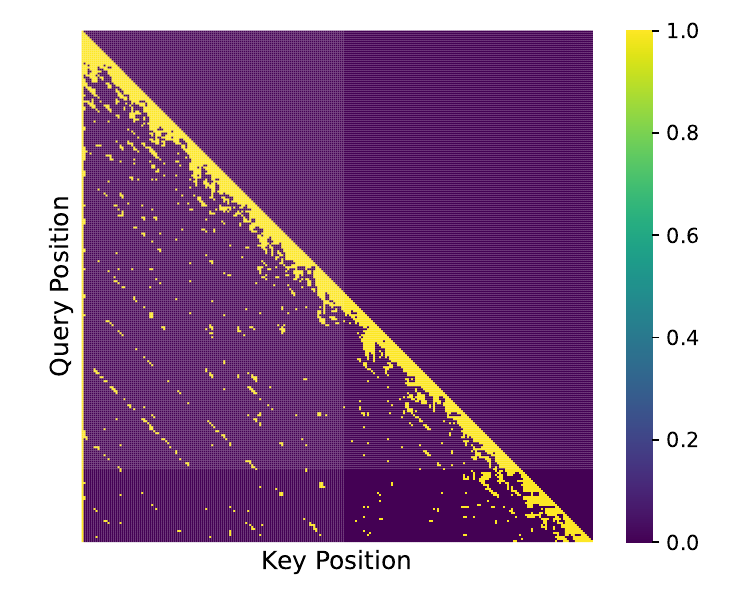}
		\end{subfigure}
		\hfill
		\begin{subfigure}[b]{0.48\textwidth}
			\includegraphics[width=\textwidth]{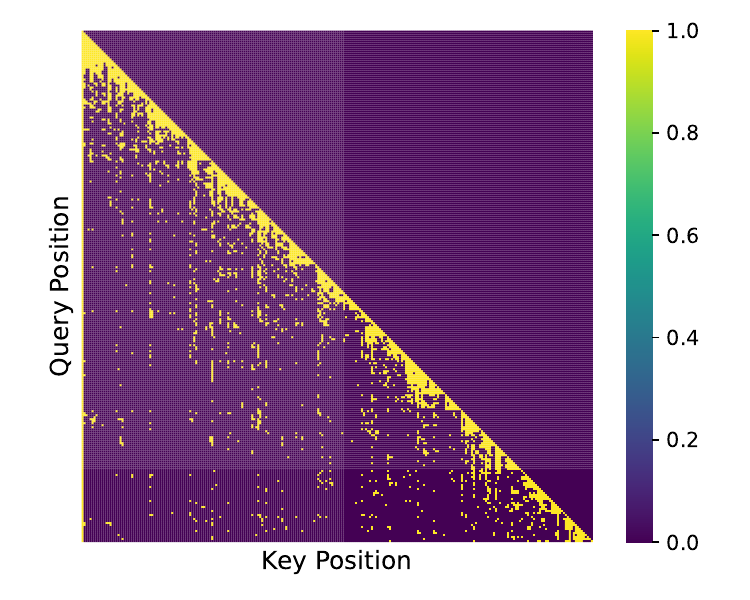}
		\end{subfigure}
		\caption{We selected the top 5\% of tokens from Llama‑3.1‑8B‑Instruct and visualized them. From this visualization, we can see that the attention patterns are not always precise.}
		\label{fig:6}
	\end{figure}

	Moreover, a detailed analysis of the attention maps reveals that the so‑called vertical and slash patterns do not continuously appear at a precise index; rather, they are intermittently distributed within a certain range around that index, as shown in Figure \ref{fig:6}. This indicates that a single original positional index cannot fully cover the effective scope of these two patterns, thereby highlighting the necessity of introducing a positional expansion indexing mechanism during decoding. Such a mechanism can capture the intermittent attention features within a wider local window, improving the model’s sensitivity to and expressiveness of cross‑interval information correlations.
	
	\subsection{Impact of Highly Sparse Heads on Pattern Capture}
	
\begin{figure}[htbp]
		\centering
		\begin{subfigure}[b]{0.3\textwidth}
			\includegraphics[width=\textwidth]{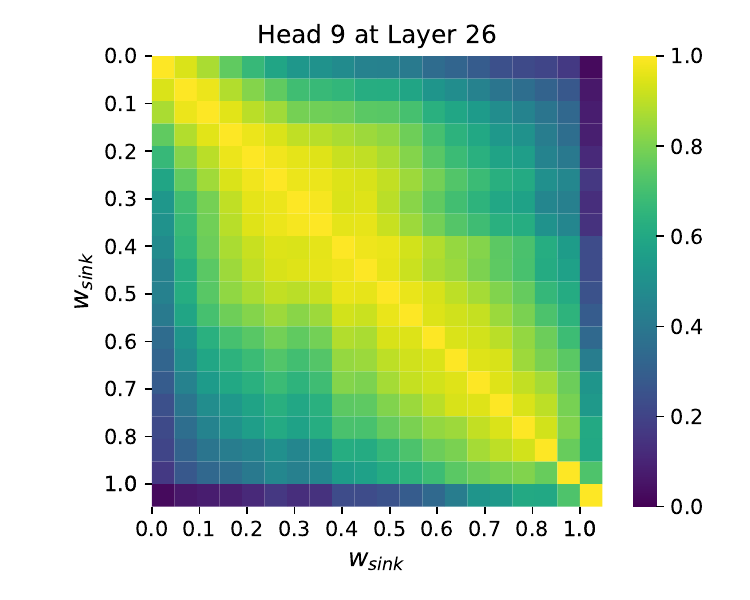}
		\end{subfigure}
		\hfill
		\begin{subfigure}[b]{0.3\textwidth}
			\includegraphics[width=\textwidth]{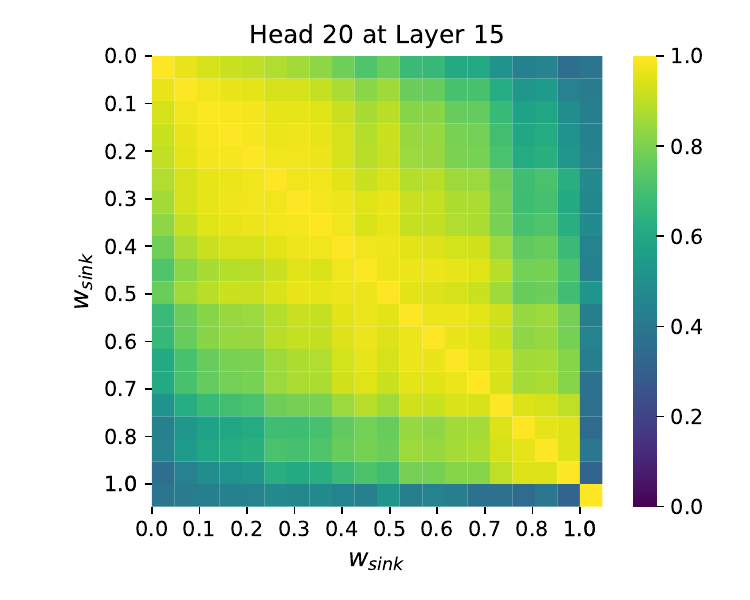}
		\end{subfigure}
		\hfill
		\begin{subfigure}[b]{0.3\textwidth}
			\includegraphics[width=\textwidth]{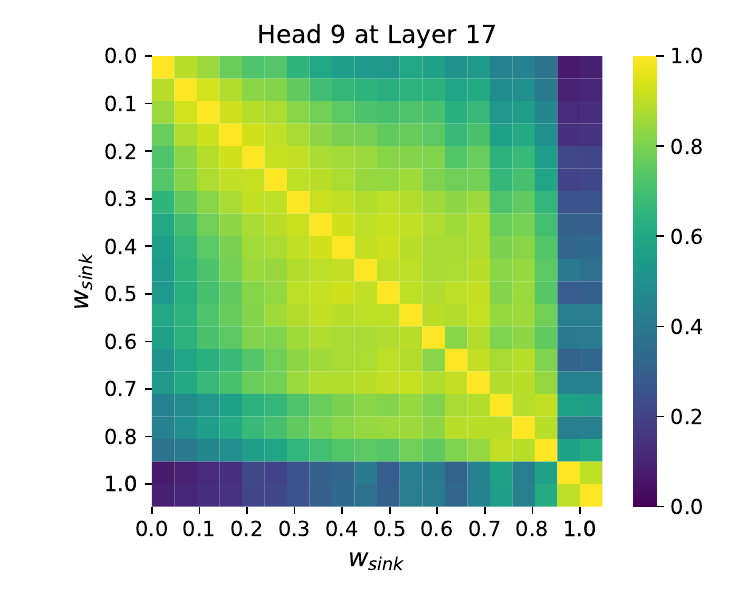}
		\end{subfigure}
		\caption{This heatmap illustrates the cosine similarity between aggregated attention patterns across multiple decoding steps, based on attention heads with varying sparsity levels. To better analyze the impact of sparsity, we divide the attention weight of $w_{sink}$
			into 5\% intervals and compute the similarity of attention patterns within each interval.}
		\label{fig:34}
	\end{figure}

	In Figure \ref{fig:34}, We group attention heads based on their sparsity levels and aggregate attention scores across different decoding steps. Subsequently, we compute the pairwise cosine similarity between the aggregated attention patterns and visualize the results as a heatmap in figure. Notably, when the attention weight assigned to the sink token exceeds 90\%, the similarity between patterns exhibits a marked shift, suggesting that sparse heads may influence the overall structure of pattern capture.
	
	\subsection{Output of Highly Sparse Heads}
	
	\begin{figure}[h]
		\centering
		\includegraphics[width=\textwidth]{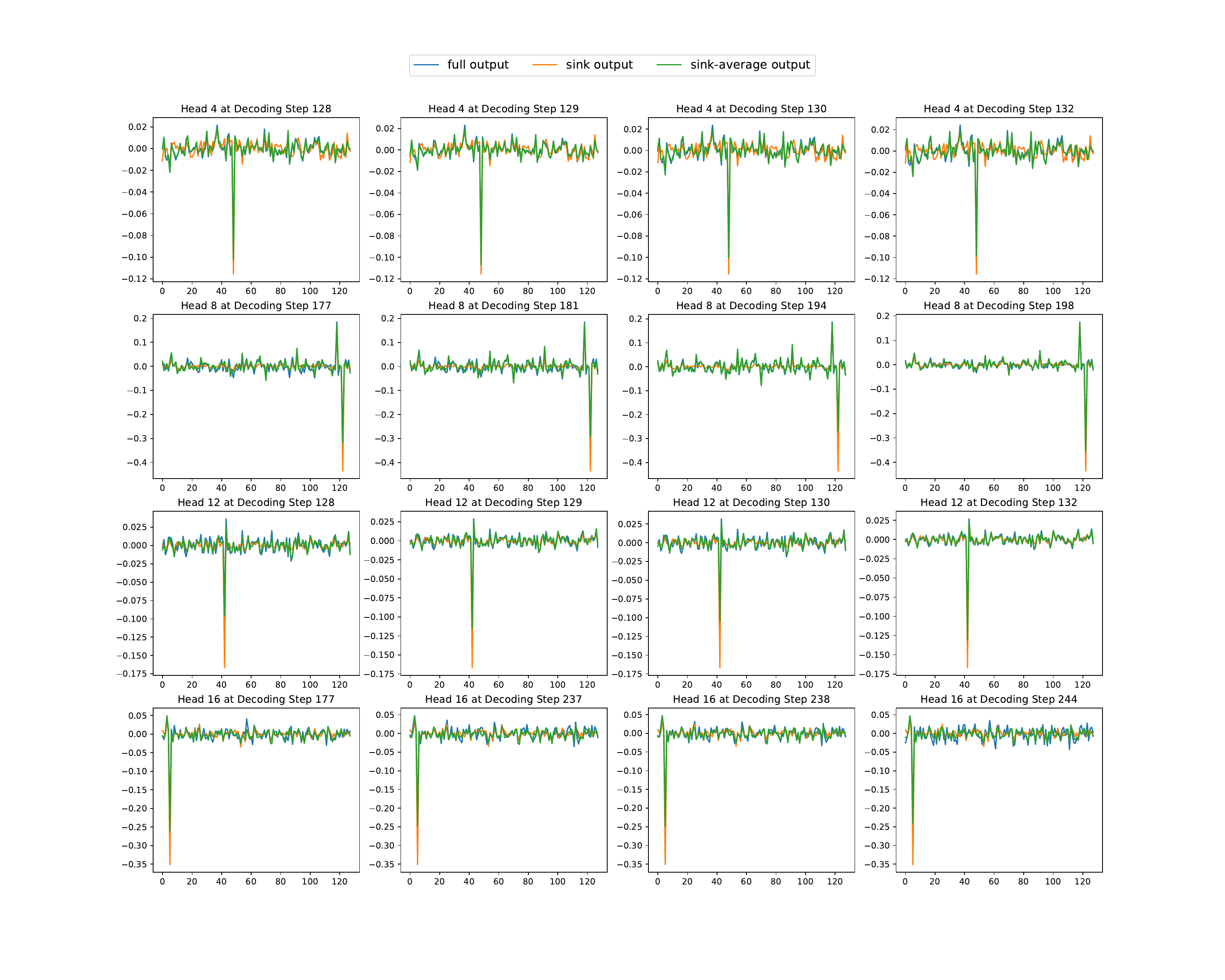}
		\caption{Visualization of output vectors from different attention heads at various decoding steps in the 23rd layer of Llama-3.1-8B-Instruct. Each output vector lies in a 128-dimensional space.}
		\label{fig:8}
	\end{figure}
	
	\section{Methodological Details}
	
	\subsection{Head sparsity estimation}
	\label{app:Head sparsity estimation}
	
	The method for computing the sparsity ratio $\rho_h^{(t)}$ in Equation~\eqref{eq:6} aims to approximate the true proportion of attention focused on the sink token by modeling how attention is distributed across three components:
	
	\begin{itemize}
		\item \textbf{Sink Attention $w^{(t)}_{sink}$}: the attention weight assigned to the sink token.
		\item \textbf{Global Attention $w^{(t)}_{global}$}: the attention distribution over all tokens (excluding the sink token). We assume that the attention weight distribution follows a log-normal distribution. Let $ X \sim \mathcal{LN}(\mu, \sigma^2) $, i.e., a log-normal distribution. By definition, this means:
		\begin{align}
			X &\sim \mathcal{LN}(\mu, \sigma^2) \Rightarrow \log X \sim \mathcal{N}(\mu, \sigma^2)
		\end{align}
		\begin{align}
			\text{Geometric Mean}[X] &= \exp\left( \mathbb{E}[\log X] \right) = \exp(\mu)
		\end{align}
		\begin{align}
			\text{Arithmetic Mean}[X] &= \exp\left( \mu + \frac{\sigma^2}{2} \right)
		\end{align}
		So, we have:
		\begin{align}
			\text{Arithmetic Mean}[X] &= \text{Geometric Mean}[X] \cdot \exp\left( \frac{\sigma^2}{2} \right)
		\end{align}
		
		\item \textbf{Local Attention $w^{(t)}_{local}$}: the sum of attention weights over the most recent tokens (e.g., the last five), reflecting sensitivity to recent context.
	\end{itemize}
	When an attention head exhibits high sparsity, its attention score distribution tends to flatten, lacking prominent peaks. In such cases, $w^{(t)}_{global}$ provides a more effective approximation of the true global attention weights. Conversely, if the attention distribution is more concentrated, the local attention term $w^{(t)}_{local}$ better captures the head’s dependency on recent context. By dynamically comparing these three metrics (sink, global, and local attention), we can selectively simplify or skip subsequent computations during decoding, thereby significantly reducing the overall computational cost.
	
	\subsection{Parameter Selection in LFPS}
	\label{app:Parameter Selection in LFPS}
	
	The decay parameter $r$ in Equation~\eqref{eq:10} controls the effective range of historical information considered during decoding. As decoding progresses, attention scores from earlier steps have diminishing influence on the current output. We set this parameter to 0.95, which results in the accumulated weight sum being approximately:
	\begin{equation}
		\sum_{i=1}^{n}\Phi_{ver}^{(t+1)}(i) = \sum_{i=1}^{n}\Phi_{sla}^{(t+1)}(i) = \frac{0.5}{1-r}
		\label{eq:12}
	\end{equation}
	Therefore, when constructing the initial score table, we choose $\frac{1}{2s(1-r)}$ as the normalization coefficient to ensure that the sum of the scores equals $\frac{1}{2(1-r)}$.

	The sparsity threshold parameter $ \varepsilon $ plays a crucial role in identifying heads that exhibit extreme concentration on the sink token during decoding. A higher value of $ \varepsilon $ enforces a stricter criterion, classifying only those heads with near-exclusive focus on the sink token as highly sparse. In our experiments, we empirically set $ \varepsilon = 0.85 $, which effectively filters out heads that could introduce instability into the attention update process due to overly dominant sink-token weights. Table~\ref{tab:sparsity_threshold} summarizes the impact of varying $ \varepsilon $ on head pruning and Qasper (8K+) accuracy.

	\begin{table}[H]
		\centering
		\caption{Impact of different sparsity thresholds $ \varepsilon $ on head pruning and Qasper(8K+)\cite{dasigi2021dataset} accuracy}
		\begin{tabular}{ccccccccc}
			\toprule
			Threshold $\varepsilon$ & 1     & 0.95  & 0.9  & 0.85  & 0.8  & 0.75 & 0.7 \\
			\midrule
			Head Cost Reduction   & 0\%   & 2.5\%   & 6.1\%  & 12\%  & 16\% & 23\% & 32\%  \\
			Qasper                & 38.94  & 39.34 & 39.21  & 38.95 & 36.65  & 36.13 & 36.29\\
			\bottomrule
		\end{tabular}
		\label{tab:sparsity_threshold}
	\end{table}

	The hyperparameter $a$ in Equation~\eqref{eq:8} controls the dynamic threshold for initial candidate selection. A larger $a$ increases the threshold
	$\tau(x)$,
	thereby reducing the size of the candidate index set and the fraction of KV pairs requiring exact dot‑product evaluation. Table~\ref{tab:a-selection} reports, for various sequence lengths, the percentage of positions for which we must perform the full Top‑$k$ inner‑product when choosing different values of $a$.
	
	\begin{table}[ht]
		\centering
		\caption{Percentage of Top‑$k$ dot‑product computations.}
		\label{tab:a-selection}
		\begin{tabular}{c|cccccc}
			$a$ & 4k & 8k & 16k & 32k & 64k & 128k \\ \hline
			0.1  & 8.0\%  & 6.2\%  & 5.0\%  & 4.3\%  & 2.6\%  & 2.0\%  \\
			0.2  & 6.0\%  & 4.6\%  & 4.1\%  & 3.4\%  & 2.0\%  & 1.7\%  \\
			0.3  & 4.8\%  & 4.1\%  & 3.7\%  & 3.0\%  & 1.7\%  & 1.4\%  \\
			0.4  & 4.2\%  & 3.6\%  & 3.2\%  & 2.5\%  & 1.5\%  & 1.2\% \\
		\end{tabular}
	\end{table}
	
\end{document}